\documentclass{article}
\usepackage{spconf,amsmath,graphicx}

\usepackage{hyperref}
\usepackage{xcolor}
\usepackage{kotex}
\usepackage{booktabs}
\usepackage{makecell}
\usepackage{multirow}
\usepackage{parskip}

\graphicspath{{figs/}}
\DeclareGraphicsExtensions{.pdf,.png,.jpg}

\newcommand{\parhead}[1]{\noindent\textbf{#1}}

\title{
Two-stage Textual Knowledge Distillation
\\ for End-to-End Spoken Language Understanding
}

\name{
Seongbin Kim$^{1,2}$\sthanks{Equal contribution.}\sthanks{SK was an intern at Clova AI while doing this work.}, 
Gyuwan Kim$^{1}$\footnotemark[1], 
Seongjin Shin$^1$, 
Sangmin Lee$^2$
}
\address{Clova AI, NAVER Corp.\textsuperscript{1} \: Dept. of Electrical and Computer Engineering, Inha University\textsuperscript{2}}

\begin{document}

\maketitle

\begin{abstract}
End-to-end approaches open a new way for more accurate and efficient spoken language understanding (SLU) systems by alleviating the drawbacks of traditional pipeline systems. Previous works exploit textual information for an SLU model via pre-training with automatic speech recognition or fine-tuning with knowledge distillation. To utilize textual information more effectively, this work proposes a two-stage textual knowledge distillation method that matches utterance-level representations and predicted logits of two modalities during pre-training and fine-tuning, sequentially. We use vq-wav2vec BERT as a speech encoder because it captures general and rich features. Furthermore, we improve the performance, especially in a low-resource scenario, with data augmentation methods by randomly masking spans of discrete audio tokens and contextualized hidden representations. Consequently, we push the state-of-the-art on the Fluent Speech Commands, achieving 99.7\% test accuracy in the full dataset setting and 99.5\% in the 10\% subset setting. Throughout the ablation studies, we empirically verify that all used methods are crucial to the final performance, providing the best practice for spoken language understanding. Code is available at \href{https://github.com/clovaai/textual-kd-slu}{https://github.com/clovaai/textual-kd-slu}.

\end{abstract}


\begin{figure}[t]
    \centering 
    \includegraphics[width=0.42\textwidth]{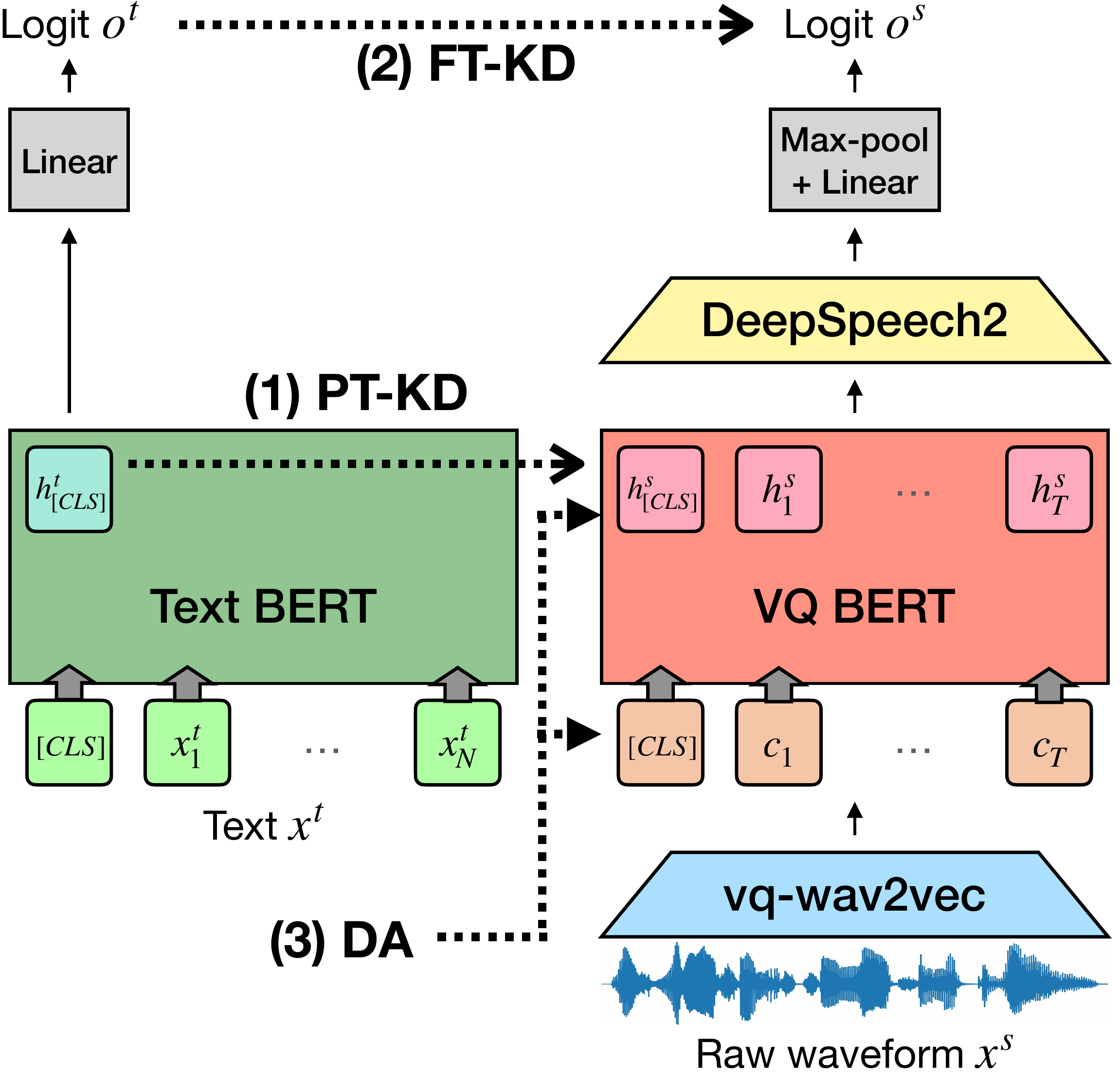}
    \caption
    {
    \label{fig:overview}
    Overview of the proposed method.
    Our E2E SLU model is a combination of vq-wav2vec BERT and DeepSpeech2 acoustic model.
    We perform knowledge distillation from the text BERT model to the speech encoder during (1) additional pre-training (PT-KD) and (2) fine-tuning (FT-KD).
    We use (3) data augmentation methods (DA) in fine-tuning.
    }
    \vskip -0.15in
\end{figure}

\section{Introduction}
\label{sec:introduction}

Conventional spoken language understanding (SLU) systems consist of the pipeline of automatic speech recognition (ASR) and natural language understanding (NLU) modules.
However, these systems are prone to error propagation because it is difficult for an NLU model to yield correct intent from erroneous ASR outputs. 
Moreover, rich information from speech signals such as prosody that might help final prediction is ignored when text inputs are given to NLU after ASR.
Therefore, end-to-end (E2E) SLU recently receives attention as a promising research direction for better SLU \cite{qian2017exploring,chen2018spoken,serdyuk2018towards}.

Despite an excellent representation capability of neural networks, a simple E2E SLU approach \cite{serdyuk2018towards} does not work well because it is difficult to extract useful features to identify intents of given utterances effectively.
This problem becomes severe when training data is scarce and noisy.
We propose a novel method based on textual knowledge distillation, pre-training, and data augmentation to address this issue.

Text is a more compressed form than audio signals to represent the same meaning. 
Therefore, a text model can guide training a speech model that requires extracting semantic representations from complex audio signals.
In this respect, we utilize knowledge distillation (KD) \cite{hinton2015distilling} to inject textual knowledge to speech encoder in both the pre-training and the fine-tuning stage.
During pre-training, we match a hidden representation of an audio sequence from a speech encoder to a hidden representation of its transcription from a text encoder.
Then, during fine-tuning, we match predicted logits for intent classification from classifiers in the two modalities.

As shown in \cite{chen2018spoken,bhosale2019end,lugosch2019speech,wang2019understanding}, pre-training speech encoder from ASR is helpful in that it learns features that are also useful for SLU to improve performance and generalize to unseen speech or text. 
Besides, we utilize vq-wav2vec BERT model \cite{baevski2019vq} as a pre-trained speech encoder.
Pre-training with self-supervised context prediction instead of only specialized for ASR can capture more general and richer representations.
For example, there might be features that can be used as a cue for SLU but not ASR. 

On top of the vq-wav2vec BERT, we use DeepSpeech2 \cite{amodei2016deep} acoustic model to aggregate feature vectors over multiple segments into a single vector for intent classification.
We pre-train this part with the ASR objective for better initialization following previous works mentioned earlier.

Motivated by SpecAugment \cite{park2019specaugment}, which is a simple and successful data augmentation technique for ASR, we apply several data augmentation methods at different positions in our model architecture and investigate their effect on SLU.
Data augmentation is also essential to achieve good performance in SLU because data augmentation has a regularization effect and is very helpful in a low-resource scenario.

In sum, we present an effective method for E2E SLU by training with knowledge distillation from the text BERT to the vq-wav2vec BERT speech encoder (see Figure~\ref{fig:overview}) and data augmentation techniques.
We empirically achieve remarkable accuracy superior to previous state-of-the-art results on the Fluent Speech Command dataset. 
Moreover, we conduct an extensive ablation study to measure the impact of each component.
Experimental results confirm that all three components are crucial to obtain the best performance: (1) textual knowledge distillation, (2) pre-training, and (3) data augmentation.

\section{Related Work}
\label{sec:related_work}

\subsection{Pre-training for SLU}

Most of the recent E2E SLU approaches consist of a two-stage procedure, pre-training a speech encoder with ASR followed by fine-tuning it for the final SLU task \cite{chen2018spoken,bhosale2019end,lugosch2019speech}.
Acoustic features learned from ASR are transferred to SLU.

Representations from self-supervised learning on a huge amount of unlabeled audio data \cite{oord2018representation,chung2019unsupervised} have shown to be effective in low-resource scenario \cite{baevski2019vq}.
We believe that they are also adequate to SLU, which usually has a few labeled data.
To the best of our knowledge, this is the first work that takes advantage of a general and powerful pre-trained speech encoder, in our case vq-wav2vec BERT, for SLU.

\subsection{Knowledge Distillation for SLU}

Several works use cross-modal distillation approach on SLU \cite{cho2020speech,denisov2020pretrained} to exploit textual knowledge.
Cho et al. \cite{cho2020speech} use knowledge distillation from a fine-tuned text BERT to an SLU model by making predicted logits for intent classification close to each other in fine-tuning.
Denisov and Vu \cite{denisov2020pretrained} match an utterance embedding and a sentence embeddings of ASR pairs using knowledge distillation as a pre-training.
Compared to them, we perform knowledge distillation in both pre-training and fine-tuning, meaning that we match sequence-level hidden representations and predicted logits of two modalities.

\subsection{Data Augmentation for SLU}

Due to the lack of high-quality SLU data, a data augmentation approach is quite essential \cite{yoo2019data}.
SpecAugment \cite{park2019specaugment,park2020specaugment} is a simple yet effective data augmentation method for E2E ASR, resulting in consistent accuracy gain.
For E2E SLU, Price \cite{price2020end} uses SpecAugment except time warping on log-spectrum input features for data augmentation.

\section{Proposed Method}
\label{sec:proposed_method}

In this section, we describe the architecture of our E2E SLU model (\textsection~\ref{sec:model_architecture}) and its training pipeline (\textsection~\ref{sec:training}).
Figure 1 illustrates an overview of our proposed method.

\subsection{Model Architecture}
\label{sec:model_architecture}

Our SLU model is a stack of vq-wav2vec BERT \cite{baevski2019vq} based speech encoder and DeepSpeech2 \cite{amodei2016deep} acoustic model (AM).
We use DeepSpeech2 to extract acoustic features but do not explicitly perform ASR.
Text is not given in the test time.
We use a text BERT model \cite{devlin2018bert} as a text encoder for textual knowledge distillation to assist the training of the SLU model.

\parhead{vq-wav2vec BERT}
vq-wav2vec maps a speech data $x^s$, which is given as a raw waveform, into a sequence of discrete codes $c = (c_1, \cdots, c_T)$.
This discrete sequence after appending a special $[CLS]$ token is represented as a sequence of contextualized hidden vectors $h^s = (h^s_{[CLS]}, h^s_1, \cdots, h^s_T)$ via a BERT model.

\parhead{DeepSpeech2 AM}
DeepSpeech2 AM consists of 2D-CNN layers and bidirectional LSTM layers.
The output of AM passes a max-pooling layer followed by a projection layer to calculate predicted logits $o^s$ for the intent classification.

\parhead{Text BERT}
A sentence $x^t = (x^t_1, \cdots, x^t_N)$ is encoded using a text BERT as a sequence of contextualized hidden vectors $h^t = (h^t_{[CLS]}, h^t_1, \cdots, h^t_N)$ after appending a special $[CLS]$ token.
$h^t_{[CLS]}$ is regarded as a sentence representation.
This sentence representation passes a projection layer to calculate predicted logits $o^t$ for the intent classification.

\subsection{Training}
\label{sec:training}

We borrow a pre-trained vq-wav2vec BERT from \cite{baevski2019vq}.
Our baseline SLU model is a combination of this vq-wav2vec BERT and DeepSpeech2 AM.
During the fine-tuning, vq-wav2vec BERT, which is regarded as a feature extractor, remains frozen.
To inject textual knowledge into the model, we perform additional training process: (1) pre-training KD, (2) fine-tuning KD, and (3) AM pre-training.

\parhead{Pre-training KD (PT-KD)}
vq-wav2vec BERT was pre-trained by the masked language modeling (MLM) task.
Using speech-text pairs from the ASR dataset, we further pre-train it by minimizing an L1 loss, $\|h^s_{[CLS]} - h^t_{[CLS]}\|_1$, to make a sequence-level contextualized representation of speech data close to that of text data in addition to the original MLM loss of \cite{baevski2019vq}.
We train MLM in conjunction with the hidden representation matching to keep learned useful acoustic features.
vq-wav2vec is not updated in additional pre-training.

\parhead{Fine-tuning KD (FT-KD)}
Compared to the SLU model given speech utterance as input, a text BERT model fine-tuned with pairs of a ground text and an intent label surprisingly performs well.
Using SLU data, we fine-tune the SLU model by minimizing an L1 loss between predicted logits of two modalities, $\|o^s - o^t\|_1$, in addition to the original supervised loss.

\parhead{AM Pre-training (AM-PT)}
Before fine-tuning the SLU model, we pre-train DeepSpeech2 AM with speech-to-text pairs for the better initialization of model weights used for SLU.
It resembles other SLU approaches that rely on ASR pre-training.
During the AM pre-training, pre-trained vq-wav2vec BERT is not updated to prevent a catastrophic forgetting of distilled weights.

\subsection{Data Augmentation (DA)}

During the fine-tuning, we apply two data augmentation methods.
First, we mask spans of consecutive discretized speech tokens similar to the pre-training of vq-wav2vec BERT \cite{baevski2019vq} though it is for the regularization effect rather than the masked prediction task.
Following SpanBERT \cite{joshi2020spanbert}, we randomly choose $p_d$ of all tokens as a starting index and mask $M_d$ consecutive tokens from every selected index with allowing overlap.
Second, we apply time masking and channel masking with the same mechanism as token masking with parameters of $(p_t, M_t)$ and $(p_c, M_c)$.
It is almost the same as SpecAugment \cite{park2019specaugment} except time-warping but differs in that masking is applied on contextualized features rather than input features.
In our preliminary experiments, we find that applying two DA methods are complementary to each other. 
Therefore, we use both of them together in our experiments.

\section{Experiments}
\label{sec:experiment_setup}

\subsection{Dataset}

We conduct experiments on three SLU datasets: Fluent Speech Commands (FSC) \cite{lugosch2019speech}, SNIPS \cite{coucke2018snips}, and Smartlights \cite{saade2018spoken}.
Table~\ref{tab:statistics} summarizes their statistics in terms of the number of speakers and utterances.

\begin{table}[t]
    \centering
    \scriptsize
    \setlength{\tabcolsep}{3pt}
    \begin{tabular}{lcccccccccccc}
    \toprule
    & \multicolumn{3}{c}{FSC (336)} & \multicolumn{3}{c}{SNIPS (7)} & \multicolumn{3}{c}{Smartlights (6)} \\
    & Train & Valid & Test & Train & Valid & Test & Train & Valid & Test \\
    \midrule
    \# Speakers   & 77 & 10 & 10 & 1 & 1 & 1 & 48 & 2 & 2 \\
    \# Utterances & 23,132 & 3,118 & 3,793 & 13,084 & 700 & 700 & 1,162 & 166 & 332 \\
    \bottomrule
    \end{tabular}
    \caption{
    Dataset Statistics of FSC, SNIPS, and Smartlights.
    They have 336, 7, and 6 intent classes, respectively.
    }
    \label{tab:statistics}
    \vskip -0.15in
\end{table}

We use FSC for comparison with other works because it is one of the most widely used datasets to evaluate SLU systems.
The class of FSC dataset is a triple of (action, object, location), so the number of total classes is 336 ($=6 \times 14 \times 4$).
Test accuracy on FSC almost reaches 100\%, implying that there is little room to improve and evaluate the newly proposed method's effectiveness.
Following \cite{lugosch2019speech,cho2020speech}, we simulate a data shortage scenario using only 10\% of the speech-text pairs in training.
We randomly divide FSC dataset into ten parts and report the average accuracy on them.

In addition to FSC, we experiment on SNIPS and Smartlights to prove that our approach is generally applicable to other settings.
SNIPS is an NLU benchmark, so they only provide text utterances and their corresponding labels.
We generate speech data by Google's commercial speech synthesis toolkit\footnote{https://cloud.google.com/text-to-speech} similar to \cite{huang2020learning} to use SNIPS for SLU evaluation.
We use a single speaker option by setting as a basic voice type named \textit{en-US-Standard-B}.
Because other works \cite{huang2020learning,cao2020style} use their own speech synthesis methods and does not mention exact details to reproduce, a fair comparison between them and ours is impossible.

Smartlights\footnote{https://github.com/sonos/spoken-language-understanding-research-datasets} \cite{saade2018spoken} has two subsets, \textit{Close} field and \textit{Far} field.
The Close set is gathered by recording utterances using crowd-sourcing.
The Far set is collected by playing these recorded utterances with a neutral speaker and recording them at a distance of 2 meters.
Those two sets include the same set of utterances, but the Far set is much noisy.
We split each set into a training, validation, and test set as the ratio of the number of utterances to be 70:10:20 following \cite{bhosale2019end}.

\subsection{Model Configurations}

We borrow a pretrained vq-wav2vec BERT supported in the fairseq \cite{ott2019fairseq} library\footnote{https://github.com/pytorch/fairseq}.
Its vector quantization is based on the k-means clustering, and its BERT is a base size of 12-layers trained similar to \cite{liu2019roberta}.
Each token after the vector quantization represents 10ms of audio data.

DeepSpeech2 AM is a stack of two 2D-CNN layers and five bidirectional LSTM layers.
The two CNN layers have a channel size of 32, kernel sizes of $41 \times 11$ and $21 \times 11$, respectively, and strides of $2 \times 2$ and $2 \times 1$, respectively.
The hidden size of all LSTM layers is 768.

We use pretrained RoBERTa-base \cite{liu2019roberta} from the fairseq for the text BERT and a softmax layer for intent classification.

\subsection{Training Details}

We further pre-train vq-wav2vec BERT from the pre-trained weights on 960h of Librispeech \cite{panayotov2015librispeech}.
The learning rate is linearly decayed from $1 \times 10^{-6}$ to 0 over 250k steps.
A slightly small learning rate is used for PT-KD because we initialize it from the pre-trained weights.
We use the batch size of 256 with 8 V100 GPUs and gradient accumulation.
To not forget general speech representations of vq-wav2vec BERT, we early stop when the MLM loss starts increasing, so training ends after the update of 10k steps. 

We fine-tune for 200 epochs with the learning rate annealed from $1 \times 10^{-4}$ by the factor of $\gamma$ at each epoch.
We use $\gamma \in \{1.01, 1.05, 1.1\}$ depending on the training set size.
For each dataset, we choose the best checkpoint in terms of validation accuracy.
We tune data augmentation hyperparameter values depending on the dataset in following ranges:
masking span length $M_d, M_c, M_t \in \{5, 10\}$ and maximum masking ratios $p_d M_d \in \{0.1, 0.2 \}$, $p_t M_t, p_c M_c \in \{0.1, 0.2, 0.3, 0.4, 0.5\}$.  

\subsection{Results and Discussion}

We do an ablation study to evaluate the effectiveness of components in our proposed method. 
Methods in the order of PT-KD, FT-KD, AM-PT, and DA starting from the baseline using our model architecture without any additional training techniques are incrementally added one-by-one.

\begin{table}[t]
    \centering
    \begin{tabular}{lrrrr}
    \toprule
    \multicolumn{1}{c}{\multirow{2}*{Method}} & \multicolumn{2}{c}{Full} & \multicolumn{2}{c}{10$\%$} \\
    \cmidrule(lr){2-3} \cmidrule(lr){4-5}
    & Valid & Test & Valid & Test  \\
    \midrule
    Lugosh et al. \cite{lugosch2019speech}          &    - & 96.6 & - & 88.9 \\
    \:\:+AM-PT \cite{lugosch2019speech}    &    - & 98.8 & - & 97.9 \\
    \:\:\:\:+FT-KD \cite{cho2020speech}         &    - & 99.0 & - & 98.1 \\
    Price \cite{price2020end}                       & 92.5 & 99.1 & - & - \\
    \:\:+DA                          & 94.4 & 99.4 & - & -  \\
    \:\:+AM-PT                            & 94.8 & 99.3 & - & - \\
    \:\:\:\:+DA                          & 96.6 & 99.5 & - & - \\
    \midrule
    \midrule
    VQ-BERT +DS2                                            & 93.1 & 98.9 & 87.3 & 97.0 \\
    \:\:+PT-KD                             & 94.1 & 99.0 & 90.7 & 98.5 \\
    \:\:\:\:+FT-KD                          & 96.2 & 99.6 & 93.3 & 99.2 \\
    \:\:\:\:\:\:+AM-PT                     & 96.4 & 99.6 & 94.3 & 99.3 \\ 
    \:\:\:\:\:\:\:\:+DA              & \textbf{97.8} & \textbf{99.7} & \textbf{96.2} & \textbf{99.5} \\
    \bottomrule
    \end{tabular}
    \caption{
    Results on FSC dataset. 
    Because test accuracy is almost close to 100$\%$, we also report validation accuracy to compare between methods.
    }
    \label{tab:fsc}
    \vskip -0.15in
\end{table}

Table~\ref{tab:fsc} display the results on FSC dataset.
Previous state-of-the-art \cite{price2020end} achieves almost perfect accuracy of 99.5\%.
Our best model surpasses that result by achieving 99.7\% test accuracy and comparable result with only 10\% of training pairs. 
Overall the gap between the full dataset setting and 10\% subset setting is reduced.
It indicates that our method is especially effective in a low-resource setting.
All of the training methods are beneficial to improve accuracy.

\begin{table}[t]
    \centering
    \begin{tabular}{lrrr}
    \toprule
    \multicolumn{1}{c}{\multirow{2}*{Method}} & \multirow{2}*{SNIPS} & \multicolumn{2}{c}{Smartlights} \\
    \cmidrule(lr){3-4} 
    & & \multicolumn{1}{c}{Close} & \multicolumn{1}{c}{Far}  \\
    \midrule
    VQ-BERT +DS2            & 86.4 & 75.9 & 47.9 \\
    \:\:+PT-KD              & 88.3 & 81.3 & 51.2\\
    \:\:\:\:+FT-KD          & 95.3 & 84.6 & 59.6 \\
    \:\:\:\:\:\:+AM-PT      & \textbf{96.7} & 92.2 & 70.5 \\ 
    \:\:\:\:\:\:\:\:+DA     & 95.7 & \textbf{95.5} & \textbf{75.0} \\
    \bottomrule
    \end{tabular}
    \caption{
    Results on SNIPS and Smartlights dataset.
    We report test accuracy.
    }
    \label{tab:snips}
    \vskip -0.15in
\end{table}

Table~\ref{tab:snips} shows the results on the SNIPS and Smartlights dataset.
We achieve fine accuracy on these datasets, although comparison with other works is difficult.
PT-KD and FT-KD boost accuracy significantly.
Data augmentation makes SLU model robust to a noisy environment and generalizes to unseen speakers.
However, data augmentation is not helpful in the SNIPS dataset.
We presume that this is because its utterances in the training set and the test set are clean and synthesized with a single speaker.

\section{Conclusion}
\label{sec:concusion}

This work proposes a novel method, which leverages textual information using knowledge distillation in both pre-training and fine-tuning stages for end-to-end spoken language understanding.
To do so, we utilize vq-wav2vec BERT as a speech encoder, which has learned general and rich speech representations and thus allows matching with text representations from the text BERT natural.
Moreover, we use data augmentation methods by randomly masking spans of representations at different positions.
With a thorough ablation study, we prove that two-stage knowledge distillation, AM pre-training, and data augmentation are crucial to learning a robust SLU model.
As a result, we achieve a state-of-the-art result on FSC and good accuracy on other datasets, SNIPS and Smartlights.
We leave the extension of our method to other downstream tasks such as speech emotion recognition \cite{siriwardhana2020jointly} and spoken question answering \cite{li2018spoken} as future work.

\section{Acknowledgements}

This research was supported by NAVER Corp.
The authors greatly appreciate Donghyun Kwak, Gichang Lee, Sang-woo Lee, Minjeong Kim, Jingu Kang, and Woomyoung Park at Naver Clova AI for constructive feedback.
We use Naver Smart Machine Learning \cite{kim2018nsml} platform for the experiments.

\clearpage

\let\OLDthebibliography\thebibliography
\renewcommand\thebibliography[1]{
  \OLDthebibliography{#1}
  \setlength{\parskip}{0pt}
  \setlength{\itemsep}{0pt plus 0.4ex}
}

\bibliographystyle{IEEEbib}
\bibliography{ms}

\end{document}